\documentclass[journal,twoside]{IEEEtran}

\ifCLASSINFOpdf
\else
   \usepackage[dvips]{graphicx}
\fi
\usepackage{url}

\usepackage{amsmath,amssymb,amsfonts}
\usepackage{graphicx}
\usepackage{bbding}
\usepackage{orcidlink}

\begin{document}

\title{SDiT: Spiking Diffusion Model with Transformer}
\author{Shu Yang~\orcidlink{0009-0008-5440-1009}, Hanzhi Ma~\orcidlink{0000-0001-7914-9323}, \IEEEmembership{Member, IEEE}, Chengting Yu~\orcidlink{0009-0007-7210-879X}, Aili Wang~\orcidlink{0000-0002-1019-4019}, \IEEEmembership{Member, IEEE}, Er-Ping Li~\orcidlink{0000-0002-5006-7399}, \IEEEmembership{Fellow, IEEE}



\thanks{This work was supported by the National Natural Science Foundation of China under Grant No.62071424 and 62027805, and National Key Research and Development Program of China under Grant No.2023YFB3812500. (Corresponding authors: Hanzhi Ma; Er-Ping Li.)}

\thanks{Shu Yang, Hanzhi Ma, Chengting Yu, Aili Wang, Er-Ping Li are with ZJU-UIUC Institute, Zhejiang University, Haining, Zhejiang 314400, China (e-mail: shu.23@intl.zju.edu.cn; mahanzhi@zju.edu.cn; chengting.21@intl.zju.edu.cn;  ailiwang@intl.zju.edu.cn; liep@zju.edu.cn).}




}

\maketitle

\begin{abstract}
Spiking neural networks (SNNs) have low power consumption and bio-interpretable characteristics, and are considered to have tremendous potential for energy-efficient computing. However, the exploration of SNNs on image generation tasks remains very limited, and a unified and effective structure for SNN-based generative models has yet to be proposed. In this paper, we explore a novel diffusion model architecture within spiking neural networks. We utilize transformer to replace the commonly used U-net structure in mainstream diffusion models. It can generate higher quality images with relatively lower computational cost and shorter sampling time. It aims to provide an empirical baseline for research of generative models based on SNNs. Experiments on MNIST, Fashion-MNIST, and CIFAR-10 datasets demonstrate that our work is highly competitive compared to existing SNN generative models.
\end{abstract}


\IEEEpeerreviewmaketitle

\section{Introduction}


\IEEEPARstart{S}{piking} neural networks (SNNs) are considered to be the third generation of neural networks with higher biological interpretability, event-driven properties, and lower power consumption, and thus have the potential to become competitive alternatives to Artificial Neural Networks (ANNs) in the future. In SNNs, all information is encoded in spike sequences, enabling SNNs to perform accumulative operations at lower power budgets for energy efficiency.

SNNs trained with deep learning techniques, especially surrogate gradient learning methods \cite{neftci2019surrogate}, have shown promising results on basic tasks like image classification and segmentation \cite{nunes2022spiking}. However, the application of SNNs on more complex computer vision tasks, especially generative models, has been limited.

Recently, diffusion models have achieved significant success on image generation \cite{dhariwal2021diffusion}, owing to their robust training objectives derived from a likelihood perspective. They provide flexibility in model designs without constraints on their network architecture  \cite{rombach2022high}. Some works \cite{peebles2023scalable}\cite{bao2023all} have attempted to use diffusion models with different backbones.
In the field of SNN-based image generation, preliminary attempts like Spiking-GAN \cite{kotariya2022spiking} and FSVAE \cite{kamata2022fully} have been made but the results are not on par with ANNs. SDDPM \cite{cao2024spiking} and SPIKING-DIFFUSION \cite{liu2023spiking} attempt to introduce diffusion models into the SNNs.

In this work, we propose Spiking Diffusion Transformer (SDiT), a novel SNN diffusion model architecture based on transformer. It demonstrates superior image generation potential for SNNs. We employ an efficient self-attention : RWKV\cite{peng2023rwkv}, and introduce the Reconstruction Module, a specially designed module aimed at supplementing information lost after the firing of spiking neurons, thereby enhancing the quality of the reconstructed image. Comprehensive experiments on MNIST \cite{lecun2010mnist}, Fashion-MNIST \cite{xiao2017fashion} and CIFAR-10 \cite{krizhevsky2009learning} show that SDiT has great competitiveness among the existing image generation models based on SNNs.

\begin{figure}[tb]
\centerline{\includegraphics[width=1.0\columnwidth]{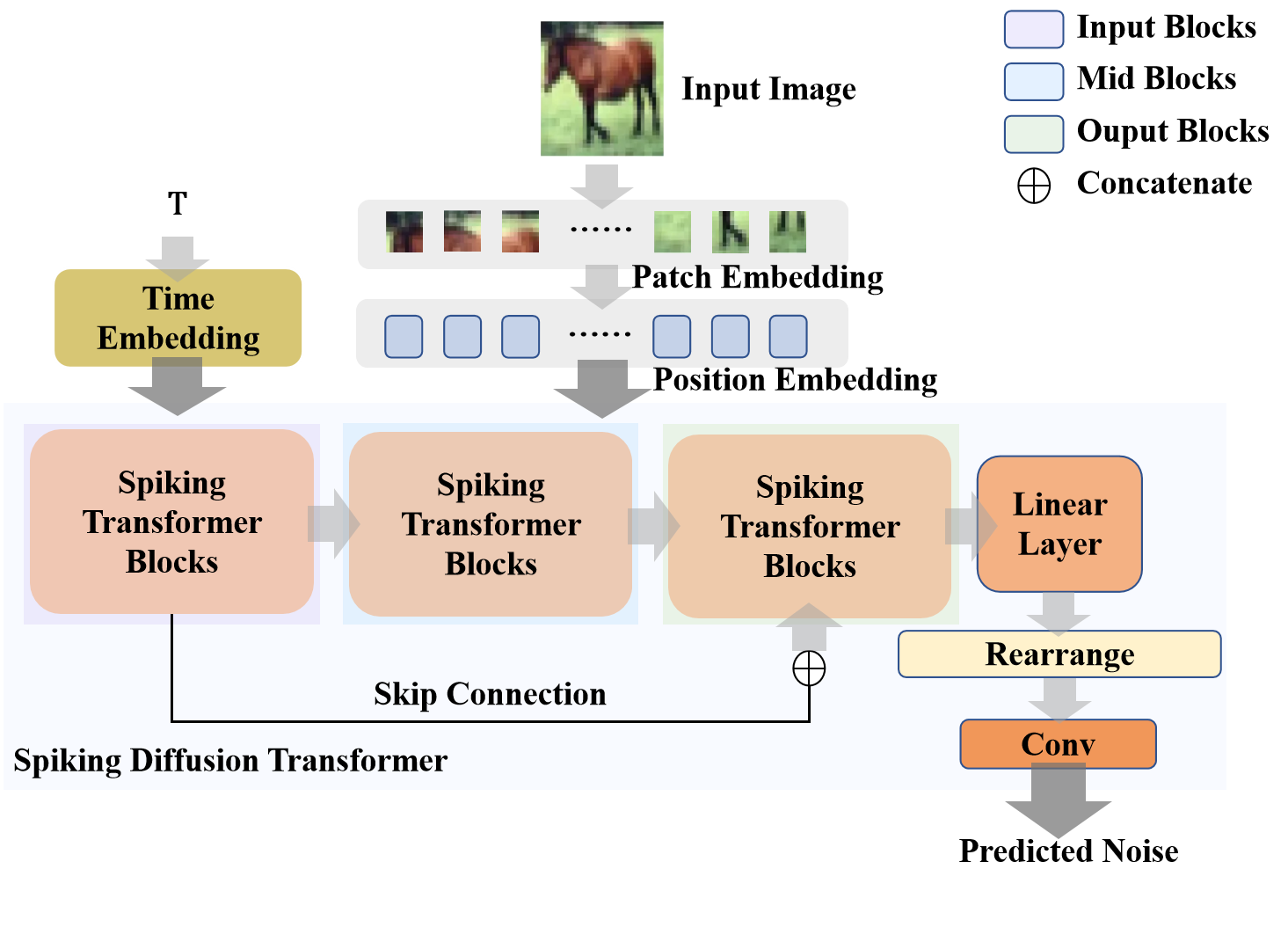}}
\caption{Diagram of SDiT architecture, illustrating the flow from input time and patch embeddings through multiple spiking transformer blocks with skip connections, culminating in a final processing stage with linear and convolutional layers for predicted noise generation.}
\label{fig1}
\end{figure}

\begin{figure}[tb]
\centerline{\includegraphics[width=1\columnwidth]{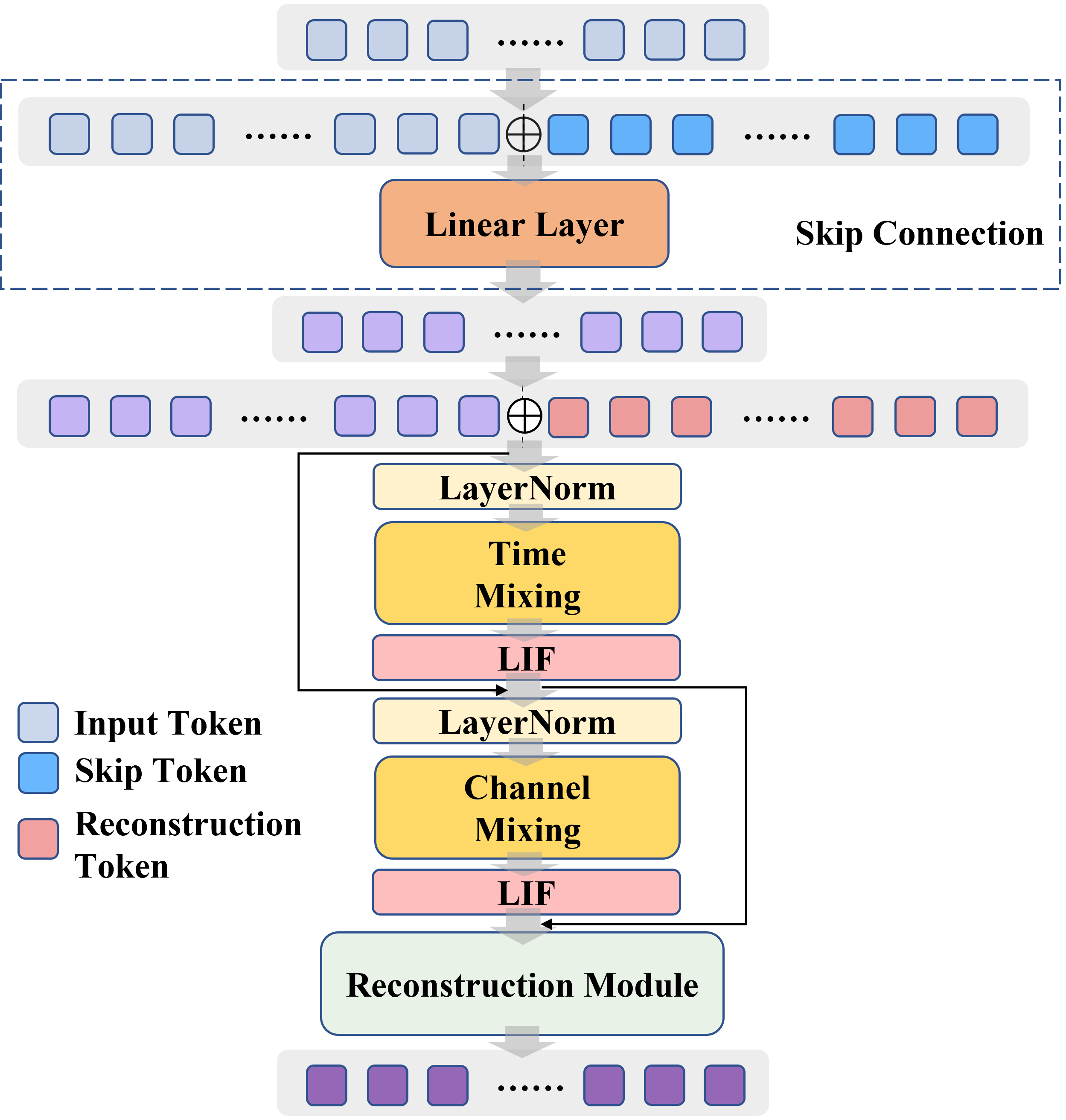}}
\caption{Spiking Transformer Block. This block comprises skip connections from the previous block, Time-Mixing, LIF neurons, Channel-Mixing, and internal residual connections. Its core structure is analogous to that of a transformer block, with the key distinction being the substitution of self-attention by RWKV. The Reconstruction Module disassembles the previously concatenated Reconstruction Tokens, supplementing the original vector with transformed information that includes the intrinsic dynamics of LIF neurons. This process compensates for the information loss in the original vector after it has been processed by spiking neurons.}
\label{fig2}
\end{figure}

\section{Preliminary}

RWKV \cite{peng2023rwkv} is a powerful competitor to the self-attention mechanism in transformer that can achieve efficient self-attention under low computational complexity.

Given an input vector $X$ at position $t$, first apply linear transformations $r_t=XW_r$, $k_t=XW_k$, $v_t=XW_v$, where $W_r$, $W_k$, $W_v$ are linear transformation matrices. $X$ is obtained by weighting the original input and input from the previous timestep, referred to as token-shift. $r_t$ is referred to as the reception matrix with each element indicating the reception of past information. $K$ and $V$ are analogous to the key and value matrices in self-attention.

The attention output of RKWV is:
\begin{equation}
wkv_t = \frac{\sum_{i=1}^{t-1}e^{-(t-1-i)w+k_i}\odot v_i+e^{u+k_t}\odot v_t}{\sum_{i=1}^{t-1}e^{-(t-1-i)w+k_i}\odot v_i}
\end{equation}
where $k_{i}$ is the vector of row $i$ in $k_t$, $v_{i}$ is the vector of row $i$ in $v_t$, $\odot$ is the Hadamard product, $w$ is a learnable position weight decay vector, and $u$ prevents the degeneration of $w$.

Time-Mixing is defined on top of the RWKV attention output by adding a linear output layer:
\begin{equation}
o_t = W_o · (\sigma(r_t) \odot wkv_t)
\end{equation}
Channel-Mixing is defined as:
\begin{equation}
o_t = \sigma(r_t) \odot (W_v · \max(k_t, 0)^2)
\end{equation}
where $W_o$ and $W_v$ are linear transformations, $\sigma$ is the Sigmoid function applied to $r_t$. By applying Sigmoid to $r_t$, unnecessary historical information is eliminated, serving as a “forget gate” to obtain the output.

Time-Mixing can be simply seen as a replacement for standard self-attention, and Channel-Mixing as a replacement for the Feed-Forward Network (FFN) layer.

\begin{figure}[tb]
    \centering
    \includegraphics[width=1\columnwidth]{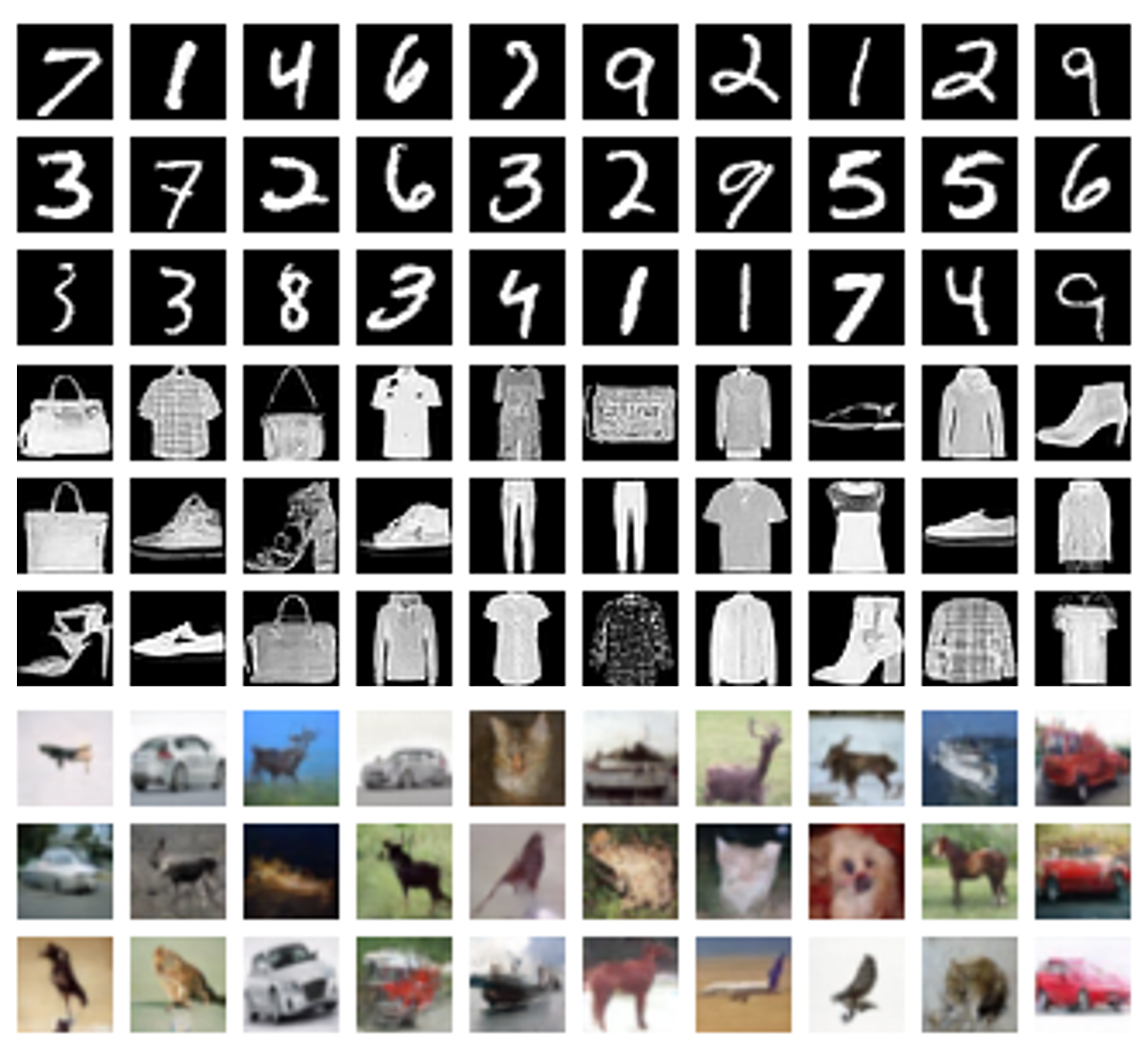}
    \caption{Samples on different datasets. The first three rows from the top represent MNIST, the middle three rows depict Fashion-MNIST, and the final three rows correspond to CIFAR-10.}
    \label{fig3}
\end{figure}
\begin{figure}[tb]
    \centering
    \includegraphics[width=1\columnwidth]{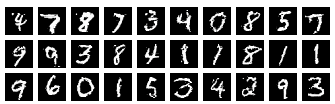}
    \caption{Samples on MNIST without Reconstruction Module.}
    \label{fig4}
\end{figure}

\section{Proposed Method}

\subsection{Overview of SDiT Architecture}




Due to the distinct neuronal mechanisms in SNNs compared to ANNs, the integration of complex self-attention mechanisms, which add to the model's parameter count, often leads to training difficulties, even when using gradient approximation techniques. To facilitate more efficient training, we propose incorporating the RWKV mechanism, known for its approximate linear computational complexity, into SNNs. This approach, inspired by the Attention Free Transformer~\cite{zhai2021attention}, simplifies the model by reducing both computational and parameter complexity. RWKV, offering performance comparable to traditional self-attention, is utilized as a more efficient alternative in our transformer model.

Fig.~\ref{fig1} shows the overview of the network architecture. Before entering the model, the input image first goes through patch embedding, which is then goes through position embedding. The denoising timestep goes through time embedding before entering the model. Inspired by the U-ViT \cite{bao2023all}, the model consists of three stages of Spiking Transformer Blocks. The Input Blocks serve as the first stage of the input processing, where each block outputs are directed into two separate paths - one that connects directly to the next block, and the other that forms skip connections to the corresponding blocks in the Output Blocks. The outputs from the Spiking Transformer Blocks are then fed into the Final Layer to be mapped back to the original image size. Following patch reconstruction and a convolution layer, the final predicted noise is obtained.

\subsection{Embedding}
Before inputting into the Spiking Diffusion Transformer, embedding operations are first performed on the image and denoising timestep.

As shown in Fig.~\ref{fig1}, the denoising timestep is embedded into the same dimensional space as the position embeddings. For the image, it is first divided into patches and mapped to a high dimensional space through patch embedding. Finally, position embedding is applied to supplement positional information for each token.

\subsection{Spiking Transformer Block}

The structure of the Spiking Transformer Block is shown in Fig.~\ref{fig2}.

Given an input vector $x \in \mathbb{R}^{B \times N \times D}$, where $B$ is the batch size, $N$ is the sequence length of patches, and $D$ is the feature dimension.

Let $\mathcal{F}$ denote the function for processing skip connections:
\begin{equation}
    x = \mathcal{F}(x,x_{skip})
\end{equation}
\begin{equation}
\mathcal{F}(x,x_{skip}) = \left\{\begin{matrix}([x;x_{skip}])W_{skip} , if \ x_{skip} \ not \ None \\x\end{matrix}\right.
\end{equation}
where $[\cdot]$ denotes concatenation, and $W_{skip}\in \mathbb{R}^{2D  \times D}$ is a linear transformation. For Input Blocks and Mid Blocks, $x_{skip}$ is $None$; for Output Blocks, $x_{skip}$ is the output from the corresponding Input Block.

Introduce a Reconstruction Token denoted as $z \in \mathbb{R}^{1 \times N \times D}$, and tiled before being concatenated to $x$:
\begin{equation}
    \hat x = [x;z\otimes \mathbf{1}_{B }]\in \mathbb{R}^{B \times 2N \times D}
\end{equation}
where $\otimes$ denotes the Kronecker product, and $\mathbf{1}_{B}$ is an all-ones vector.

The concatenated $\hat{x}$ is then fed into the Time-Mixing and Channel-Mixing modules. After passing through Leaky Integrate-and-Fire (LIF) neurons, residual connections are added:
\begin{equation}
    x_{Attn}  = \hat{x}  + LIF(TimeMixing(LN(\hat{x})))
\end{equation}
\begin{equation}
    x_{FFN} = x_{Attn}  + LIF(ChannelMixing(LN(x_{Attn})))
\end{equation}
where $LN$ denotes $LayerNorm$ \cite{ba2016layer}.

Finally, the Channel-Mixing output is processed:
\begin{equation}
y = \mathcal{G}(x_{FFN})
\end{equation}
where $\mathcal{G}$ denotes operations on the Reconstruction Module.

\subsection{Reconstruction Module}

Time-Mixing is a unique self-attention mechanism. Directly applying standard self-attention spaces with complex representations to the spiking representations in SNNs leads to significant information loss when the outputs are input into spiking neurons. 

To compensate for this, we designed the Reconstruction Token and Reconstruction Module. The Reconstruction Token is comprised of learnable parameters. After embedding the Reconstruction Token, it can represent the intrinsic dynamic information of spiking neurons. In the Reconstruction Module, the original outputs are rescaled to mitigate the information loss when passed through spiking neurons. The effectiveness of this is shown in Section~\ref{section5}.

After computations over the concatenated vector, the embedded Reconstruction Token is separated from the output vector.
\begin{equation}
    y',z' = \text{split}(x_{FFN})
\end{equation}
where $y' \in \mathbb{R}^{B \times N \times D}$, $z' \in \mathbb{R}^{B \times N \times D}$, and $\text{split}$ denotes the vector separation operation.

Since the Reconstruction Token has the same dimensions as the input, its feature dimension is first linearly transformed to match the number of patches:
\begin{equation}
    z'_{D} = z' W_{D} \in \mathbb{R}^{B \times N \times N}
\end{equation}
where $W_D \in \mathbb{R}^{D \times N}$, transforming along the second dimension.

Swapping the patch and feature dimensions, the patch dimension is then linearly transformed again to match the feature dimension:
\begin{equation}
    z'_{N} = z'_{D} W_{N} \in \mathbb{R}^{B \times N \times D}
\end{equation}
where $W_N \in \mathbb{R}^{N \times D}$, transforming along the second dimension.

Finally, the obtained vector is element-wise multiplied with the separated vector and added to it as supplementary information:
\begin{equation}
    y = y' + z'_{N}\odot y'
\end{equation}
where $\odot$ denotes the Hadamard product.

\begin{table}[t]
\caption{Results on Different Datasets.}
\label{tab1}
\begin{center}
\begin{tabular}{ccccc}
\hline
\textbf{Dataset}   & \textbf{Model}   & \textbf{Time Steps}   & \textbf{FID} &\textbf{IS}   \\ \hline
        &    SGAD       &     16         &  69.64   &    -    \\
        &    FSVAE      &     16         &  97.06   &   6.209     \\ 
MNIST   &    Spiking-Diffusion       &    16        &    37.50   &   -     \\
        &    SDDPM      &    4          &   29.48  &    -    \\
        &    \textbf{SDiT}   &  4           &  \textbf{5.54}   &    2.452    \\ \hline 
        &    SGAD       &     16         & 165.42    &    -    \\
        &    FSVAE       &     16         &   90.12  &   4.551     \\ 
Fashion-MNIST   &    Spiking-Diffusion       &     16       &    91.98   &   -     \\
        &    SDDPM      &     4         &  21.38   &    -    \\
        &    \textbf{SDiT}       &  4            & \textbf{5.49}    &    4.549    \\ \hline 
        &    SGAD       &       16       &  181.50   &    -    \\
        &    FSVAE       &      16        &   175.50   &    2.945    \\ 
CIFAR-10 &    Spiking-Diffusion       &    16        &    120.50   &    -    \\
        &    SDDPM      &     4        &   16.89   &    7.655    \\
        &     \textbf{SDiT}      &     4         &  \textbf{22.17}   &   4.080     \\ \hline
\end{tabular}
\end{center}
\end{table}

\begin{table}[t]
\caption{Ablation Experiments Evaluating FID on Different Datasets.}
\label{tab2}
\begin{center}
\begin{tabular}{cc|c|cc}
\hline
\textbf{Model} &    & \textbf{Reconstruction Module} &\textbf{MNIST}  & \textbf{Fashion-MNIST}   \\ \hline
SDiT    &    &  \XSolidBrush  &224.66  &  111.52     \\
SDiT &    & \CheckmarkBold &5.54   &   5.49  \\ \hline 

\end{tabular}
\end{center}
\end{table}

\begin{table}[t]
\caption{Number of Parameters and MAC Operations.}
\label{tab3}
\begin{center}
\begin{tabular}{c|c|c|c}
\hline
\textbf{Model}& \textbf{Backbone}    & \textbf{Parameters} &\textbf{MACs}    \\ \hline
DiT& Transformer       &  13.86 M &   3.48 G      \\
SDiT &RWKV &     11.67 M & 1.32 G   \\ \hline 
\end{tabular}
\end{center}
\end{table}

\subsection{Final Layer}

The main function of this layer is to convert the output to a reconstructed image. After the Spiking Transformer Blocks, a linear transformation along the feature dimension is applied to match the original image size. Afterwards, the patches are reconstructed back into the image. Finally, a 3×3 convolutional block is applied based on \cite{bao2023all}, which can enhance the quality of generated samples.

\section{Experiment}\label{section5}

\subsection{Experiment Settings}


\subsubsection{Evaluation Metrics}

We use Fr\'echet  Inception Distance (FID) \cite{heusel2017gans} to assess the quality of sampled images and Inception Score (IS) \cite{salimans2016improved} to evaluate sample diversity. When computing FID, 50,000 images are sampled from the dataset and 50,000 images are generated to calculate FID between them. When computing the IS for MNIST and Fashion-MNIST datasets, the quantitative comparisons of IS to scores on ImageNet hold limited meaningfulness, owing to the substantial domain discrepancy between these data distributions versus ImageNet on which IS was originally proposed and validated.

\subsubsection{Implementation details}
We standardized the input size to 28×28 for both MNIST, Fashion-MNIST, and resized CIFAR-10 images for uniformity. The architecture for MNIST and Fashion-MNIST involves 2 Input and Output Blocks, 1 Mid Block, and a hidden dimension of 384. For CIFAR-10, this extends to 4 Input and Output Blocks, 1 Mid Block, and a 512 hidden dimension. Optimization is performed using AdamW with a learning rate of 1e-4, training MNIST and Fashion-MNIST models for 1600 epochs, and CIFAR-10 for 2000 epochs post-resizing. Our SNN models are developed using the SpikingJelly framework \cite{fang2023spikingjelly}. Experiments were conducted on 4×NVIDIA 4090 GPUs.





\subsection{Comparisons}

As depicted in Table~\ref{tab1}, our SDiT outperforms SGAD, FSVAE, Spiking-Diffusion and SDDPM on image reconstruction and generation on MNIST and Fashion-MNIST datasets. On CIFAR-10, the FID score of our SDiT is slightly higher than SDDPM but substantially lower than other methods, exhibiting strong image generation capabilities. Fig.~\ref{fig3} shows example of generated images from our method on the three datasets. It can be observed that under very short spiking time steps, our approach can synthesize sharper images with more defined edges.

\subsection{Ablation Study}\label{Ablation Study}

If the transformer structure is directly introduced into SNNs without adaptation, there would be immense information loss compared to the sparse spiking signals emitted by SNN neurons. To address this problem, we designed the Reconstruction Module to align both.

As shown in Table~\ref{tab2}, models without the Reconstruction Module have very high FID scores on MNIST and Fashion-MNIST datasets. The generated image quality is also poor, with blurred edges and noise dominating the scene, as shown in Fig.~\ref{fig4}. In contrast, our model with the Reconstruction Module encodes the intrinsic dynamics of SNN neurons through the evolution of Reconstruction Tokens in the data flow. This leads to significantly better performance.

Additionally, our method employs RKWV as an alternative self-attention which effectively integrates with SNNs. When configured with the same model settings, as depicted in Table~\ref{tab3}, our approach attains a smaller parameter size and lower multiply–accumulates (MACs) compared to the pure transformer ANN implementation (DiT)~\cite{peebles2023scalable}, thus fully demonstrating the low-power advantage of SNNs.






\section{Discussion}

Despite the relatively strong performance, SDiT has some limitations. The performance of SDiT on MNIST and Fashion-MNIST significantly surpasses other methods. However, on CIFAR-10, its results fail to reach state-of-the-art levels. We attribute this underperformance, to some extent, to the inferior efficacy of Vision transformers when trained on limited data. While images in MNIST and Fashion-MNIST encode relatively simple visual information, the intrinsic complexity and comparatively low resolution of CIFAR-10 images obscures fine-grained edge details in SDiT's learned representations. Regarding the integration of self-attention mechanisms and SNNs, we plan to incorporate more fine-grained schemes in future work to further minimize the information loss of self-attention within the SNN framework.

Another limitation is that the embedding component of SDiT is not implemented with a pure SNN. The current SDiT consists of a hybrid architecture of ANN and SNN. As image sizes scale up, the parameter and computation amounts in the ANN module also increase, which may diminish the inherent low-power characteristics of SNNs due to parameter growth in the portion of ANN. In future work, our plan is to explore designs fully contained within the SNN framework and attempt to construct models with a pure SNN architecture.

\section{Conclusion}

In this work, we propose a novel generative spiking neural network architecture : SDiT, which combines the low power consumption property of SNNs with superior generative capabilities. As a new effort that incorporates transformer as the backbone for diffusion models in the SNN domain, SDiT has advanced the state-of-the-art in terms of generated image quality and exhibits strong performance across multiple datasets. We envision that this research can introduce new perspectives to the field of SNN generative model for future developments in this area.

\nocite{*}
\bibliographystyle{ieeetr}
\bibliography{bib}

\end{document}